\documentclass{article}
\usepackage{placeins}

\usepackage{microtype}
\usepackage{graphicx}
\usepackage{subcaption}
\usepackage{booktabs} 

\usepackage{hyperref}

\usepackage[accepted]{icml2026}

\usepackage{amsmath}
\usepackage{amssymb}
\usepackage{mathtools}
\usepackage{amsthm}

\usepackage{url}
\usepackage{booktabs} 
\usepackage[normalem]{ulem}
\useunder{\uline}{\ul}{}
\usepackage{amsmath} 
\usepackage{listings}
\usepackage{wrapfig}
\usepackage{colortbl}
\usepackage{multirow}
\usepackage{caption}
\usepackage{subcaption}
\usepackage{xcolor}

\usepackage[breakable, most]{tcolorbox}
\definecolor{myframecolor}{HTML}{8babbe}
\definecolor{mycolback}{HTML}{e6e6e6}

\definecolor{Gray}{gray}{0.9}
\definecolor{deepgreen}{HTML}{00CC00}

\definecolor{New}{HTML}{000000}
\newcommand{\new}[1]{\textcolor{New}{#1}}

\newcommand{\tightcolorbox}[2]{%
  \begingroup
  \setlength{\fboxsep}{0.5pt}
  \colorbox[HTML]{#1}{%
    \raisebox{0pt}[1.0\ht\strutbox][1.0\dp\strutbox]{#2}%
  }%
  \endgroup
}

\theoremstyle{plain}

\theoremstyle{definition}

\theoremstyle{remark}

\usepackage[textsize=tiny]{todonotes}

\icmltitlerunning{Inside the Visual Mind: Neuroscience-Motivated Concept Circuits for Interpreting and Steering Vision Transformers}

\begin{document}

\twocolumn[
  \icmltitle{Inside the Visual Mind: Neuroscience-Motivated Concept Circuits \\ for Interpreting and Steering Vision Transformers}



  \icmlsetsymbol{equal}{*}

  \begin{icmlauthorlist}
    \icmlauthor{Tang Li}{yyy}
    \icmlauthor{Yanlin Chen}{yyy}
    \icmlauthor{Mengmeng Ma}{xxx}
    \icmlauthor{Xi Peng}{xxx}
  \end{icmlauthorlist}

  \icmlaffiliation{yyy}{Department of Computer \& Information Science, University of Delaware, Newark, DE, USA}
  \icmlaffiliation{xxx}{Department of Computer Science, University of Virginia, Charlottesville, VA, USA}

  \icmlcorrespondingauthor{Tang Li}{tangli@udel.edu}
  \icmlcorrespondingauthor{Xi Peng}{naq5rd@virginia.edu}

  \icmlkeywords{Machine Learning, ICML}

  \vskip 0.3in
]



\printAffiliationsAndNotice{}  

\begin{abstract}
Despite high accuracy, Vision Transformer (ViT) predictions can be driven by spurious cues, raising the need to understand their inner workings before safe deployment. Sparse autoencoders (SAEs) provide a promising lens for decomposing model representations into human-interpretable concepts, yet adapting SAE-based interpretation to ViTs remains challenging due to limited control over concept coverage and subjective, non-scalable feature interpretation. To fill the gaps, motivated by neuroscience-inspired principles, we propose ViSAE, a mechanistic interpretability toolbox for understanding ViT inner workings through concept circuits. ViSAE consists of three components: (1) A probing suite with 64K images and a 16K visually grounded concept vocabulary, improving concept coverage efficiency by 20× over ImageNet and interpretation accuracy by 28.7\% over existing concept sets. (2) Top-down concept reading and Bottom-up circuit tracing algorithms that automatically recover ViT inner workings via concept circuits. (3) Applications for auditing and steering ViT behavior. Through concept editing, ViSAE improves the worst-group accuracy on WaterBirds by 48.2\%, outperforming existing methods by 23.8\%. Our data and code: \url{https://github.com/deep-real/ViSAE}.
\end{abstract}
\section{Introduction}
\label{sec:introduction}

\begin{figure}[t]
 \centering
  \includegraphics[width=0.42\textwidth]{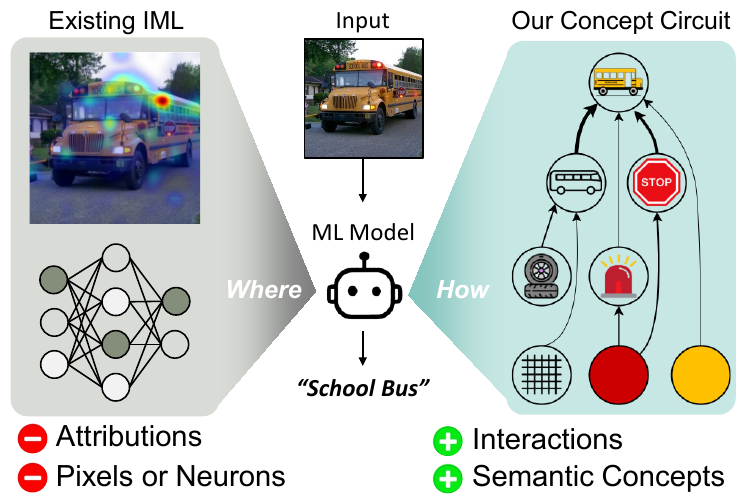}
  \caption{
  Existing interpretable machine learning methods (IML) mainly identify {\it where} the evidence is, while our concept circuits reveal {\it how} concepts interact across layers to support a prediction.
  }
  \label{fig:title}
\end{figure}

Machine learning (ML) models, such as {\it Vision Transformers (ViTs)}~\citep{dosovitskiy2020image}, have become ubiquitous foundations of high-impact systems.
Despite their strong empirical performance, their internal mechanisms remain opaque to users, revealing little about how information is represented, transformed, and used inside the network.
This opacity makes it difficult to diagnose failures, identify unsafe reasoning patterns, or intervene when a model relies on brittle or spurious features~\citep{arjovsky2019invariant, sagawadistributionally}.
Conventional {\it interpretable machine learning (IML)} methods (Fig.~\ref{fig:title}) often explain ML models by attributing predictions to input features~\citep{selvaraju2017grad, lundberg2017unified} or internal neurons~\citep{ghorbani2020neuron, bau2017network}.
However, attribution-based correlations provide limited insight into the model's underlying computation and decision process~\citep{olah2020zoom}.
A natural question arises:
{\it Can we ``read the mind'' of a large vision model?
Or, can we understand the inner workings of vision models (layer by layer, neuron by neuron, from input to output) using human-understandable concepts?}



Recent advances in {\it Mechanistic Interpretability (MI)} provide a promising foundation for this goal, especially in language Transformers.
A typical approach is to train \emph{Sparse Autoencoders (SAEs)}, which decompose polysemantic, internal representations into more monosemantic, human-interpretable concept features~\citep{bricken2023towards, huben2023sparse, zou2023representation}.
However, directly porting SAE-based workflows from language to vision runs into two practical bottlenecks.
\textbf{(1) Concept coverage lacks control.}
Existing practice~\citep{rao2024discover, stevens2025sparse, thasarathan2025universal} often trains SAEs using large generic datasets such as ImageNet~\citep{deng2009imagenet}.
This imposes limited control over which concepts are covered and how densely each abstraction level is sampled.
As a result, the concepts learned by SAEs are often biased toward dominant dataset content, such as objects (Tab.~\ref{tab:image_set}), narrowing their interpretability to deep layer semantics and leaving low- and mid-layer concepts underrepresented.
\textbf{(2) Interpretation is hard to scale.}
Unlike language features, the vision features learned by SAEs are often not naturally interpretable to humans.
A common workflow retrieves the top-activating images for each SAE feature and summarizes them into concept labels~\citep{lim2024sparse, pach2025sparse}. However, such summarizations can be subjective, summarizer-dependent, and hard to scale across tens of thousands of SAE features.

To address these gaps and enable holistic analysis of ViT inner workings, we propose \textbf{ViSAE}, a comprehensive mechanistic interpretability toolbox. ViSAE integrates probing data for SAE training and interpretation, algorithms for concept reading and circuit tracing, and practical applications for model auditing and steering.
Critically, its design is motivated by neuroscience-inspired principles for studying biological vision and neural computation.
Specifically, our toolbox answers three research questions:

\textbf{(1) Data: How can we improve concept coverage for SAE training and interpretation?}
To interpret ViTs layer by layer, SAE training requires probing examples that cover the full spectrum of visual processing.
Motivated by the hierarchical organization of the human visual system~\citep{goodale1992separate, carandini2005we}, we organize visual concepts into four abstraction levels: Primitive, Intermediate, Object, and Scene.
To train SAEs, we curate 64K probing examples from seven vision datasets, selected so that each example’s primary content aligns with one of these levels.
To interpret SAEs, we construct a 16K vocabulary grounded in our images and spanning the same hierarchy (e.g., ``stripes'', ``skimming'').
Our design improves concept coverage efficiency by $20\times$ compared with the standard ImageNet baseline.
Moreover, when using our concept set to interpret the same SAEs, we achieve a 28.7\% gain in interpretation accuracy over existing concept sets.

\textbf{(2) Algorithm: How can we reveal the inner workings of ViTs?}
While SAEs extract discrete concepts from internal representations, ViT mechanisms depend not only on which concepts are present, but also on how concepts interact across layers (Fig.~\ref{fig:title}).
This aligns with two complementary views in cognitive science for understanding biological intelligence: the Hopfieldian view emphasizes transformations in representational spaces, while the Sherringtonian view emphasizes connections among neural units and structured pathways~\citep{barack2021two}.
Motivated by these views, we propose a two-stage tracing algorithm.
{\it (i) Top-down concept reading:} trains an SAE at each Transformer layer and maps learned features to our concept vocabulary through the vision-language embedding space of CLIP~\citep{radford2021learning}, avoiding manual summarization.
\textit{(ii) Bottom-up circuit tracing:} estimates cross-layer causal effects through counterfactual interventions, producing a directed interaction graph of concepts.
Our interpretations faithfully capture model behavior, outperforming existing counterparts by 23.8\% in downstream steering tasks.

\textbf{(3) Applications: How can concept circuits improve trust in ViTs?}
Our toolbox enables diagnostic auditing and corrective steering of ViTs.
For {\it auditing,} it enables users to trace internal decision-making processes of the model, localize the visual evidence of concepts on pixels, and diagnose and summarize model failure modes.
More importantly, for {\it steering,} it offers a set of conceptual ``knobs'' to precisely control model behavior by editing concepts within representations.
For example, by turning down spuriously correlated concepts ({\it e.g.,} land backgrounds), our method improves the worst-group accuracy on the WaterBirds~\citep{sagawadistributionally} dataset by 48.2\%.

\textbf{Our Contributions:}
Unlike most existing efforts that propose another SAE variant, our ViSAE toolbox takes a {\it data-centric} perspective, providing the missing {\it infrastructure} needed to train and interpret SAEs for holistic analysis of ViT inner workings.
Our toolbox consists of:
(1) A neuroscience-motivated probing suite (64K images, 16K concepts) for SAE training and auto-interpretation.
(2) A two-stage causal tracing algorithm for layer-wise discovery of concept circuits within ViTs.
(3) Extensive empirical validation, including SAE benchmarking and applications in representation auditing and steering.
\section{Method}
\begin{figure*}[t]
  \centering
   \includegraphics[width=1.0\linewidth]{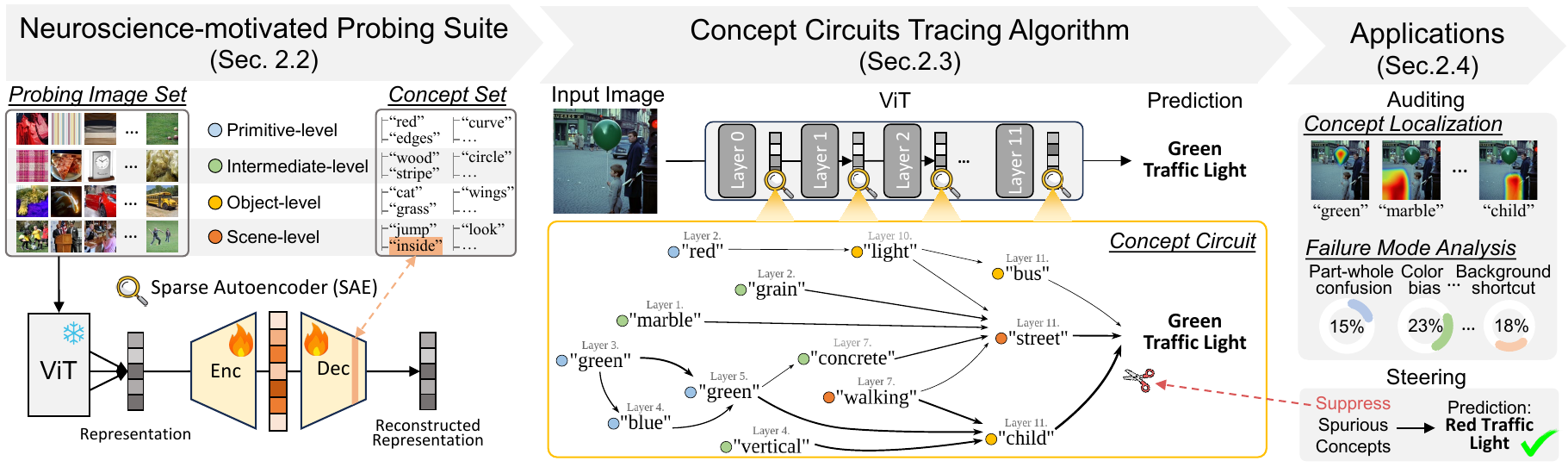}
   \caption{
   Overview of our {\it ViSAE} toolbox for interpreting ViT inner workings.
   {\it \textbf{Left:}} Motivated by the human visual cortex hierarchy, we construct a probing suite (64K images + 16K concepts) for SAE training and interpretation.
   {\it \textbf{Middle:}} Our top-down concept reading and bottom-up concept circuit tracing algorithms.
   {\it \textbf{Right:}} Our mechanistic view of ViT inner workings enables various downstream applications, such as concept localization, failure mode analysis, and model steering.
   }
   \label{fig:overview}
\end{figure*}

In this section, we first review the basics of SAEs (Sec.~\ref{subsec:sae_prelim}), then introduce our probing suite (Sec.~\ref{subsec:probing}), followed by how it is used to train SAEs and to trace concept circuits (Sec.~\ref{sec:method:tracing}), and finally how to use the toolbox to audit and steer ViTs (Sec.~\ref{sec:method:toolkits}).
See overview in Fig.~\ref{fig:overview}.

\subsection{Preliminary}
\label{subsec:sae_prelim}

Sparse Autoencoders (SAEs)~\citep{ng2011sparse, bricken2023towards} were proposed to interpret polysemantic neurons by formulating it as a {\it sparse dictionary learning} problem~\citep{olshausen1997sparse}.
The objective is to learn an {\it overcomplete} set of sparse, disentangled basis features ({\it i.e.,} concepts) that can reconstruct the input data through linear combination~\citep{thasarathan2025universal}.
Concretely, an SAE consists of an encoder that expands the input dimensionality, namely $f:\mathbb{R}^{d} \to \mathbb{R}^{m}$, $m >  d$, and a decoder $g:\mathbb{R}^{m} \to \mathbb{R}^{d}$.
A vanilla ReLU-SAE is given by:
\begin{equation}
\begin{aligned}
&\mathbf{h} = f(\mathbf{x}) = \mathrm{ReLU}(\mathbf{W}_\mathrm{enc} \mathbf{x} + \mathbf{b}_\mathrm{enc}),
\\
&\hat{\mathbf{x}} = g(\mathbf{h}) = \mathbf{W}_\mathrm{dec}\mathbf{h} + \mathbf{b}_\mathrm{dec},    
\end{aligned}
\end{equation}
where $\mathbf{W}_\mathrm{enc}, \mathbf{W}_\mathrm{dec}^{\top} \in  \mathbb{R}^{m\times d}$ and $\mathbf{b}_\mathrm{enc}, \mathbf{b}_\mathrm{dec} \in \mathbb{R}^{m}$.
Note that in the decoder parameter matrix $\mathbf{W}_\mathrm{dec}$, each column $\mathbf{w}_i$ represents a learned basis feature, namely $\hat{\mathbf{x}} = \sum_{i=0}^{m-1} h_i \mathbf{w}_i + \mathbf{b}_\mathrm{dec}$.
Its training objective minimizes the reconstruction error while enforcing sparsity in latent code:
\begin{equation}
\mathcal{L}(\mathbf{x}) = ||\mathbf{x} - \hat{\mathbf{x}}||_2^2 + \lambda ||\mathbf{h}||_1.
\end{equation}
Given the SAE, two factors largely affect interpretation quality.
{\bf (1) Quality of the input representation $\mathbf{x}$.}
The data distribution ({\it e.g.,} images) that produces $\mathbf{x}$ governs what features are even learnable.
As noted in Sec.~\ref{sec:introduction}, object-centric datasets bias learning toward specific abstraction levels~\citep{stevens2025sparse, thasarathan2025universal}.
{\bf (2) Interpretation method of SAE features.} 
Existing works typically label a feature $\mathbf{w}_i$ by inspecting its top-activating images~\citep{thasarathan2025universal, pach2025sparse}, but this process is subjective and hard to scale.
We address these challenges in the following sections with a new probing suite and an automated interpretation method.


\subsection{Neuroscience-Motivated Probing Suite}
\label{subsec:probing}
As discussed in Sec.~\ref{sec:introduction}, for SAE training, object-centric datasets often provide limited concept coverage for the full spectrum of visual processing.
To fill this gap, we construct a probing suite that offers broad concept coverage motivated by the hierarchical organization of the human visual cortex from neuroscience~\citep{goodale1992separate, carandini2005we, dicarlo2012does}.


\begin{figure}
  \begin{center}
  \includegraphics[width=0.38\textwidth]{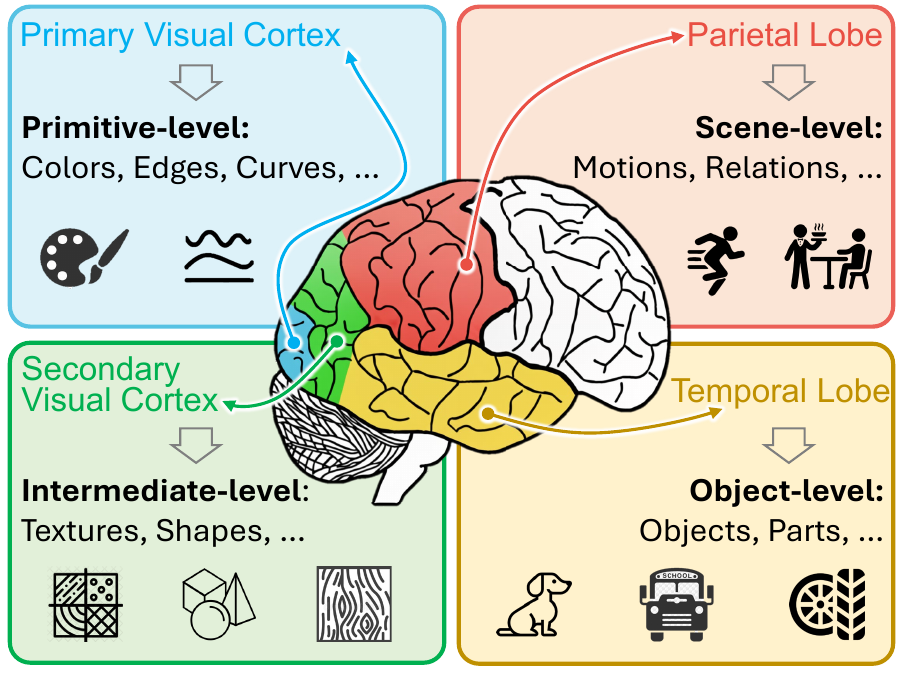}
  \end{center}
  \vspace{-5pt}
  \caption{
  \new{Our design mirrors the hierarchy of human visual cortex.}
  }
\label{fig:visual_pathway}
\end{figure}

\textbf{Background: visual cortex hierarchy.}
The human visual system processes information along abstraction levels.
As shown in Fig.~\ref{fig:visual_pathway}:
(1) At the \tightcolorbox{9FDCEF}{\it primitive level}, the primary visual cortex encodes basic visual primitives, {\it e.g.,} colors, edges, and curves.
(2) At the \tightcolorbox{C5E0B4}{\it intermediate level}, secondary visual cortex integrates these primitives into more complex patterns {\it e.g.,} textures, materials, and geometric shapes.
(3) At the \tightcolorbox{FFD966}{\it object level}, these patterns are combined into identifiable entities like tables, airplanes, or animals, supporting object recognition in the temporal lobe.
(4) Finally, at the \tightcolorbox{EB786F}{\it scene level}, higher-order regions represent actions, spatial relations, and interactions, enabling reasoning about context and events, often associated with the parietal lobe.

\begin{table*}[t]
\centering
\caption{
    \new{Comparison of concept coverage (\%) across probing image sets.
    We obtain the concept coverage of images by calculating the CLIP embedding similarity between images and our ground truth concepts (details in {\bf Appendix~\ref{appendix:concept_coverage}}).
    As shown, our probing image set demonstrates superior concept coverage across all levels of visual abstraction, with over 20$\times$ higher coverage efficiency.}
}
\resizebox{1.0\linewidth}{!}{%
\new{\begin{tabular}{ccccccclc}
\toprule[1pt]
\multirow{2}{*}{Probing Image Set} & \multirow{2}{*}{Data Source}                                                                                                                                                                                                                                                                               & \multirow{2}{*}{\# of Images} & \multicolumn{5}{c}{Concepts Covered by Images (\%)}                           & \multirow{2}{*}{\begin{tabular}[c]{@{}c@{}}Coverage Efficiency $\uparrow$\\ (\%/1K Images)\end{tabular}} \\ \cmidrule{4-8}
                                   &                                                                                                                                                                                                                                                                                                            &                               & Primitive     & Intermediate  & Object        & Scene         & Avg.          &                                                                                                          \\ \midrule[0.75pt]
ImageNet                           & ImageNet                                                                                                                                                                                                                                                                                                   & 1,281K                        & {\ul 81.0}    & {\ul 78.2}    & \textbf{97.7} & 59.0          & {\ul 78.9}    & 0.06                                                                                                     \\
MSCOCO                             & MSCOCO                                                                                                                                                                                                                                                                                                     & 118K                          & 69.6          & 65.4          & 80.4          & \textbf{63.1} & 69.6          & {\ul 0.59}                                                                                               \\
\rowcolor{Gray} Ours               & \multicolumn{1}{l}{\begin{tabular}[c]{@{}l@{}}\tightcolorbox{9FDCEF}{Primitive Level:} DTD, Broden;\\ \tightcolorbox{C5E0B4}{Intermediate Level:} Broden, ShapeNet;\\ \tightcolorbox{FFD966}{Object Level:} ImageNet, VisualGenome;\\ \tightcolorbox{EB786F}{Scene Level:} Place365, MSCOCO;\end{tabular}} & 64K                           & \textbf{87.1} & \textbf{80.6} & {\ul 92.6}    & {\ul 61.7}    & \textbf{80.5} & \textbf{1.26}                                                                                            \\ \bottomrule[1pt]
\end{tabular}}
}
\label{tab:image_set}
\end{table*}

\textbf{Construct probing image set.}
To maximize the concept coverage of SAE, we first collect probing images from seven vision datasets mirroring the hierarchy above.
Specifically, (1) at the {\it primitive level,} we collect images from the DTD~\citep{cimpoi14describing} and Broden~\citep{bau2017network};
(2) at the {\it intermediate level}, we collect images from Broden and ShapeNet~\citep{chang2015shapenet};
(3) at the {\it object level}, we collect images from ImageNet~\citep{deng2009imagenet} and Visual Genome~\citep{krishna2017visual};
and (4) at the {\it scene level}, we collect images from Places365~\citep{zhou2017places} and MSCOCO~\citep{lin2014microsoft}.\
However, naively aggregating images reduces SAE training efficiency while offering minimal concept coverage benefits.
This is because repeated views cause SAEs to waste limited model capacity on the same high-frequency concepts, biasing the learned features.
To address this issue, we prune the initial pool (121K raw image candidates) by removing one image from every pair with a cosine similarity greater than 0.85 in the {\it CLIP-ViT-B-32} embedding space.
As a result, our final set contains 64K probing images.
\new{Tab.~\ref{tab:image_set} shows our superior concept coverage, outperforming the popular ImageNet baseline by 20$\times$ in coverage efficiency.}

\textbf{Construct concept set.}
To enable the proposed automatic interpretation (Sec.~\ref{sec:method:tracing}) of SAE features, a candidate pool of concepts with strong visual grounding is required.
Existing vocabularies are typically mined from text, {\it e.g.,} frequent n-grams from Google Books~\citep{oikarinen2022clip} or LAION captions~\citep{bhalla2024interpreting}, which are typically skewed toward linguistically frequent terms and drift from the images.
To reduce this bias, we generate concepts \emph{from the images themselves}. Concretely, for each image in our probing set, we use GPT-5~\citep{openai_gpt5_system_card_2025} to annotate present concepts under the same four-level hierarchy.
The resulting concept set contains 16K unique one- and two-gram concepts (Tab.~\ref{tab:concept_set}).
Compared with existing concept sets, our concepts are 20.6\% less redundant and 26.2\% more visually grounded (Tab.~\ref{tab:concept_quality}), and it outperforms existing concept counterparts in interpretation accuracy (Tab.~\ref{tab:interpret_acc}; See details in Sec.~\ref{subsec:interp_acc}).

\textbf{Human Evaluations.}
To ensure the quality of the concept annotations, we further conduct human evaluations.
We evaluate two metrics: Faithfulness, which measures whether concepts are visually grounded in the image, and Comprehensiveness, which measures whether they cover all visual abstraction levels.
We randomly sample three groups of images (3$\times$50) paired with our concept annotations and hire graduate students to evaluate all different groups on a 0-5 Likert scale. 
As shown in Tab.~\ref{tab:concept_set}, our concepts are of high quality and the annotation process is consistent across images, as evidenced by an average rating above 4.7/5.

\begin{table}[t]
\centering
\caption{Concept set statistics and human evaluation results.}
\label{tab:concept_set}
\resizebox{1.0\linewidth}{!}{%
  \begin{minipage}{\linewidth}
    \centering
    \begin{minipage}[t]{0.38\linewidth}
      \centering
      \resizebox{\linewidth}{!}{%
        \begin{tabular}{lr}
          \toprule[1pt]
          Level        & \# of Concepts \\ \midrule[0.75pt]
          Primitive    & 1,073          \\
          Intermediate & 1,723          \\
          Object       & 10,534         \\
          Scene        & 2,720          \\ \midrule[0.5pt]
          Total        & 16,050         \\
          \bottomrule[1pt]
        \end{tabular}
      }
    \end{minipage}
    \begin{minipage}[t]{0.6\linewidth}
      \centering
      \resizebox{\linewidth}{!}{%
        \new{\begin{tabular}{ccc}
          \toprule[0.75pt]
          Evaluator & Faithfulness    & Completeness    \\ \midrule[0.5pt]
          Human\_A  & 4.65 $\pm$ 0.06 & 4.72 $\pm$ 0.04 \\
          Human\_B  & 4.83 $\pm$ 0.20 & 4.73 $\pm$ 0.06 \\
          Human\_C  & 4.90 $\pm$ 0.05 & 4.78 $\pm$ 0.18 \\ \midrule[0.5pt]
          Avg.      & 4.79            & 4.74            \\ \bottomrule[0.75pt]
        \end{tabular}}
      }
    \end{minipage}
  \end{minipage}
}
\end{table}
\begin{table}[t]
\centering
\caption{Comparison of concept set quality. Redundancy: the proportion of concept pairs with CLIP text-embedding cosine similarity $>$0.8. Visually grounded: the proportion of concept-image pairs with CLIP cosine similarity $>$0.3.}
\resizebox{\linewidth}{!}{%
\begin{tabular}{cccc}
\toprule[0.75pt]
Concept Set & \# of Concepts & Redundancy ($\downarrow$) & Visually Grounded ($\uparrow$) \\ \midrule[0.5pt]
LAION-freq  & 15K            & 0.419                           & 0.602                                \\
Google-freq & 20K            & 0.654                           & 0.432                                \\
LaBo        & 10K            & 0.584                           & 0.565                                \\
\rowcolor{Gray}Ours        & 16K            & \textbf{0.213}                  & \textbf{0.864}                       \\ \bottomrule[0.75pt]
\end{tabular}
}
\label{tab:concept_quality}
\end{table}


\subsection{Concept Circuit Tracing Algorithm}
\label{sec:method:tracing}

\textbf{Top-down concept reading.}
Although SAEs decompose polysemantic representations into disentangled features, interpreting the semantics of these features remains challenging.
Existing methods typically retrieve top-activating samples and summarize them into a specific concept~\citep{bills2023language, pach2025sparse}.
However, for vision models, this process relies on subjective human review and does not scale.
To address this issue, we introduce an automated method to ``read'' concepts directly from representations.
Specifically, by leveraging the aligned embedding space of vision-language models, we map each SAE feature to the most semantically aligned textual concept in our concept set.
Let $\mathbf{W}_\mathrm{dec}$ be the decoder weight matrix of a trained SAE, where each column $\mathbf{w}_i$ is a basis feature.
Using our probing image set $\mathcal{D}_\text{probe} = \left \{ x_1, ..., x_N \right \}$, we extract the feature activation vector $q_i = \big[\, h_i(x_1), h_i(x_2), \dots, h_i(x_N) \,\big]^\top \in \mathbb{R}^N$ for neuron $i$ over all images.
For the concept set $\mathcal{D}_\text{concept} = \left \{ c_1, ... , c_M \right \}$, we compute a concept activation matrix $P \in \mathbb{R}^{N \times M}$ using a VLM, {\it e.g.,} CLIP~\citep{radford2021learning}, where $P_{nm}$ is the embedding similarity between image $x_n$ and concept $c_m$.
We associate SAE feature $i$ with concept $c_m$ using the Soft Weighted Point-wise Mutual Information (Soft-WPMI)~\citep{oikarinen2022clip}:
\begin{equation}
\resizebox{0.91\linewidth}{!}{$
\text{Sim}(i, c_m) =\log \,\mathbb{E}_{x \sim \mathcal{D}_{\text{probe}}}\!\left[ \alpha_i(x) \cdot P_{xm} \right] - \lambda \log p(c_m),
$}
\end{equation}
where $\alpha_i(x_n) = \frac{\exp(q_i[n])}{\sum_{j=1}^N \exp(q_i[j])}$ is the softmax-normalized activation of neuron $i$, $p(c_m) = \frac{1}{N} \sum_{n=1}^N P_{nm}$ is the marginal prevalence of concept $c_m$, and $\lambda > 0$ controls the penalty for overly frequent concepts.
The final concept label for SAE feature $i$ is determined by:
\begin{equation}
c^*(i) = \arg\max_{c_m \in \mathcal{D}_\text{concept}} \text{Sim}(i, c_m).
\end{equation}

\textbf{Bottom-up circuit tracing.}
Although we have mapped SAE features to concepts, how these discrete concepts relate to one another and how they compose into the final decision remains unclear. 
To address this, we trace the causal influence among concepts and toward the prediction from bottom up.
Specifically, inspired by {\it activation patching} methods~\citep{meng2022locating, conmy2023towards}, we define edges by quantifying the causal importance via counterfactual interventions.
Let $\alpha_j^t$ be the activation of a target concept $c_j^t$ in a downstream
target layer $t$.
Consider a concept $c_i^s$ extracted from source layer $s$ via SAE, with activation $\alpha_i^s$.
To measure its influence on a target concept $c_j^t$, we construct two layer $s$ representations of the same input $x$:
the original representation $r_{\text{clean}}$ and a \emph{patched} representation $r_{\text{patch}}$, in which $c_i^s$ is ablated by setting its activation $\alpha_i^s$ to zero and reconstructing the representation via the SAE decoder.
In this case, we define the causal influence of $c_i^s$ on $c_j^t$ by measuring the \emph{interventional effect} (IE)~\citep{pearl2001direct}:
\begin{equation}
\begin{aligned}
\hspace{-5pt}
\text{IE}_{i \to j}^{s \to t} &= \text{IE}\!\left(\alpha_j^t; c_i^s; r_{\text{clean}}, r_{\text{patch}}\right)\\
&= \alpha_j^t\!\left(r_{\text{clean}}\right)
- \alpha_j^t\!\left(r_{\text{clean}} \,\middle|\, \text{do}\!\left(\alpha_i^s = \alpha_i^s(r_{\text{patch}})\right)\right).
\end{aligned}
\end{equation}
We can also obtain the contribution of concept $c_i^s$ to the final prediction $y$ by:
\begin{equation}
\begin{aligned}
\hspace{-15pt}
\text{IE}_{i \to y}^{s}
&= \text{IE}\!\left(y; c_i^s; r_{\text{clean}}, r_{\text{patch}}\right)\\
&= y(x_{\text{clean}})
- y\!\left(r_{\text{clean}} \,\middle|\, 
   \text{do}\!\left(\alpha_i^s = \alpha_i^s(r_{\text{patch}})\right)\right). 
\end{aligned}
\end{equation}
By repeating this procedure for all concepts over all layers, we obtain a directed graph where nodes represent concepts and edges are weighted by the causal importance of the target node regarding the final prediction.
Concretely, the weight of an edge from $c_i^s$ to $c_j^t$ is given by $\text{IE}_{i \to j}^{s \to t} \cdot \text{IE}_{j \to y}^{t}$.
\new{We model the ViT forward pass as a deterministic structural causal model (SCM), and our concept nodes refine this into a more fine-grained SCM on top of it.
Consequently, all edges are directed and acyclic, {\it i.e.,} $s \to t$ only when $t > s$.}
This \emph{concept circuit} reveals how primitive features are progressively composed into intermediate patterns and, ultimately, high-level semantics that drive the model’s predictions.

\subsection{Applications}
\label{sec:method:toolkits}

Building upon the concept circuits, our {\it ViSAE} toolbox offers a practical toolkit for model analysis and intervention.

\textbf{Auditing.}
\textit{(1) Trace information flow:}
Users can visualize the concept circuit for an arbitrary image, revealing the pathways of causal influence from low-level primitives to the final prediction (Fig.~\ref{fig:information_flow}).
\textit{(2) Localize concepts on pixels:}
Concretely, to localize $c_i$ on images, we calculate the cosine similarities between $\mathbf{w}_i$ and each image token $t_j \in \mathbb{R}^{d}$ from the corresponding transformer layer.
This generates a saliency map $\textstyle h_i = \frac{\left\langle  \mathbf{w}_i, t_j \right \rangle}{\|\mathbf{w}_i\|_2 \; \|t_j\|_2}$ that highlights the regions where the concept $c_i$ is most strongly activated (Fig.~\ref{fig:localization}).
\textit{(3) Diagnose failure modes:}
By comparing concept circuits from correct and incorrect predictions, users can systematically diagnose failure modes and determine whether errors arise from spurious cues or missing critical factors (Fig.~\ref{fig:failure_mode}).

\textbf{Steering.}
Our {\it ViSAE} enables targeted control of model behavior by concept editing.
\textit{(1) Suppress spurious concepts:}
Mitigate shortcut learning by setting the activation of an undesired concept to zero, effectively removing its influence from the computation graph.
\textit{(2) Amplify robust concepts:}
Enhance the effect of desirable features by manually increasing their activation (Tab.~\ref{tab:steering}).
\section{Experiments}

\begin{figure*}[t]
  \centering
   \includegraphics[width=0.85\linewidth]{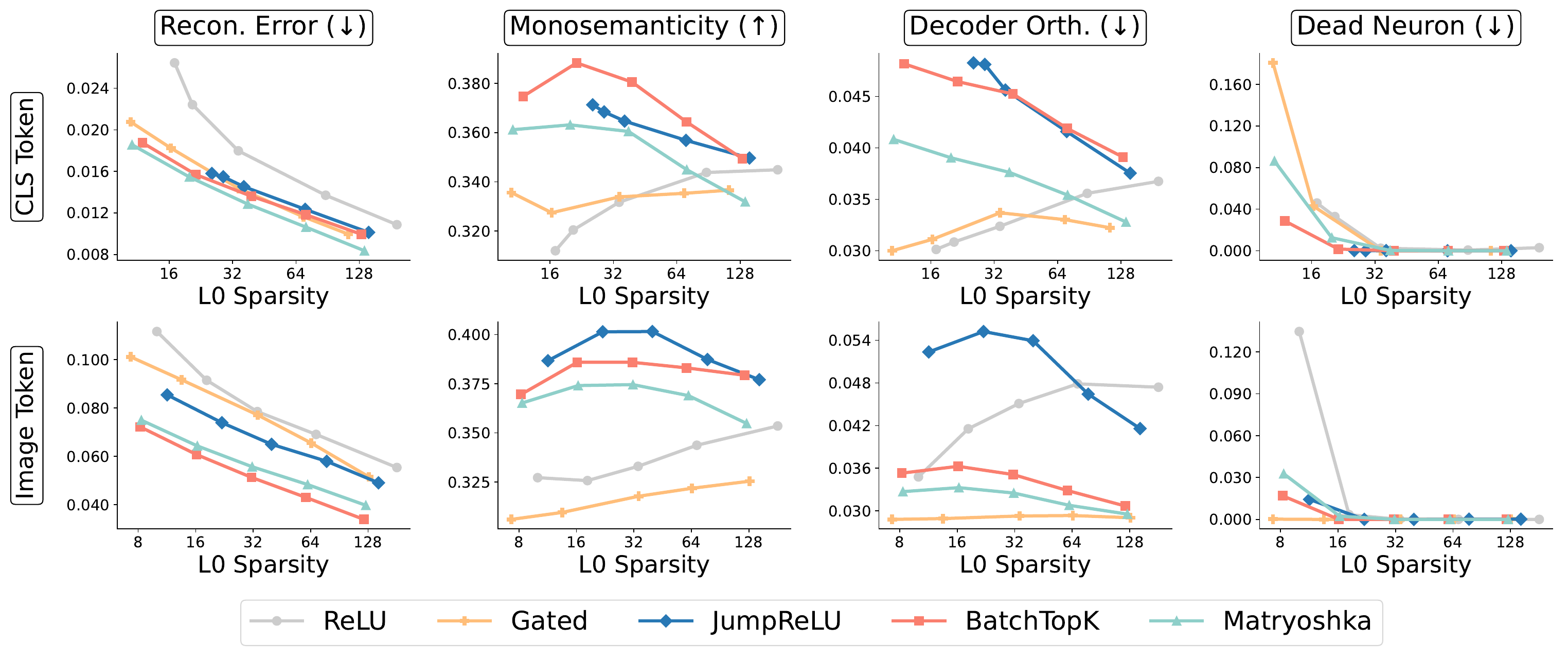}
   \vspace{-8pt}
   \caption{
   Benchmark results for expansion factor ($8\times$).
   As shown, BatchTopK-SAE strikes a better trade-off across all metrics.
   Therefore, in subsequent experiments we use the BatchTopK-SAE with expansion factor = $8\times$ and $L_0$ Sparsity = 128.
   Full tables in \textbf{Appendix~\ref{appendix:benchmark}.}
   }
   \vspace{-5pt}
   \label{fig:benchmark}
\end{figure*}

\subsection{Benchmarking SAEs on Image Data}
\label{subsec:benchmark}
We systematically evaluate SAE variants on disentangling and reconstructing vision representations.

\textbf{Settings:}
We evaluate representative SAEs, {\it i.e.,} ReLU-SAE~\citep{bricken2023towards}, BatchTopK-SAE~\citep{bussmann2024batchtopk}, Matryoshka-SAE~\citep{bussmann2025learning}, Gated-SAE~\citep{rajamanoharan2024improving}, JumpReLU-SAE~\citep{rajamanoharan2024jumping}.
For each SAE architecture, we train multiple variants with five expansion factors ($2\times$, $4\times$, $8\times$, $16\times$ and $32\times$) and five average $L_0$ sparsities ($8$, $16$, $32$, $64$ and $128$), resulting in 25 instances in total.
The {\it expansion factor} refers to the expansion ratio between the SAE latent dimension and input dimension.
We train SAEs on the representations (CLS and image token separately, 2$\times$12 SAEs in total) extracted from the residual stream of each layer in ViT ({\it CLIP-ViT-B-32} unless noted), using our probing image set as input.
For each layer, we train SAEs separately on {\it cls} tokens and {\it image} tokens.
Details of these SAE architectures are in \textbf{Appendix~\ref{appendix:SAE}}.

\textbf{Metrics:}
{\it (1) $L_0$ Sparsity}: the average number of non-zero activations.
{\it (2) Reconstruction Error (RE)}: the mean squared error (MSE) between the input representation and the SAE reconstructed representation.
{\it (3) Decoder Orthogonality (DO)}~\citep{zaigrajew2025interpreting}: the mean cosine similarity between each pair of decoder columns. This metric measures whether an SAE encodes concepts with distinct semantic meanings.
{\it (4) Dead Neuron (DN)}: the proportion of SAE basis features remaining consistently inactive (zero activation) across the whole training dataset.
{\it (5) Monosemanticity (MS)}~\citep{pach2025sparse}: whether each basis feature of an SAE consistently activates on images of the same semantics.
Details of the metrics are in \textbf{Appendix~\ref{appendix:evaluation_metircs}.}

\textbf{Results:}
We observe that all SAE architectures share similar trends with respect to $L_0$ sparsity on all five evaluation metrics.
To balance the reconstruction quality and monosemanticity, we set the sparsity to $128$ and the expansion factors to $8\times$ in practice.
Fig.~\ref{fig:benchmark} shows the benchmark results for expansion factor $8\times$.
Implementation details and full tables are in \textbf{Appendix~\ref{appendix:implementation_details}~\&~\ref{appendix:benchmark}.}


\begin{table} 
\centering
\caption{Comparison of Interpretation Accuracy.}
\vspace{-5pt}
\resizebox{0.97\columnwidth}{!}{%
\begin{tabular}{ccccc}
\toprule[1pt]
\multirow{2}{*}{\begin{tabular}[c]{@{}c@{}}Probing Image\\ Set\end{tabular}} & \multirow{2}{*}{\begin{tabular}[c]{@{}c@{}}Concept\\ Set\end{tabular}} & \multicolumn{3}{c}{Interpretation Accuracy (\%)} \\ \cmidrule{3-5} 
                                                                             &                                                                        & Top-10         & Top-20         & Top-30         \\ \midrule[0.75pt]
Broden                                                                       & Broden                                                                 & 16.8           & 23.9           & 26.8           \\ \midrule[0.5pt]
\multirow{4}{*}{ImageNet}                                                    & LaBo-ImageNet                                                          & 15.9           & 18.4           & 20.3           \\
                                                                             & Google-20K                                                             & 7.3            & 10.2           & 12.0           \\
                                                                             & LAION-15K                                                              & 16.5           & 21.7           & 24.0           \\
                                                                             & Ours-16K                                                               & 32.2           & 43.5           & 50.7           \\ \midrule[0.5pt]
\multirow{3}{*}{MSCOCO}                                                      & Google-20K                                                             & 10.0           & 13.3           & 15.3           \\
                                                                             & LAION-15K                                                              & 16.3           & 20.4           & 23.7           \\
                                                                             & Ours-16K                                                               & 34.9           & 45.0           & 51.2           \\ \midrule[0.5pt]
\multirow{3}{*}{Ours-64K}                                                    & Google-20K                                                             & 10.7           & 14.4           & 16.7           \\
                                                                             & LAION-15K                                                              & 17.3           & 22.2           & 25.4           \\
                                                                             \rowcolor{Gray}& Ours-16K                                                               & \textbf{36.6}  & \textbf{47.6}  & \textbf{54.1}  \\ \bottomrule[1pt]
\end{tabular}
}
\label{tab:interpret_acc}
\end{table}

\begin{table}[t]
\centering
\caption{Comparison with MLLM-based Summarization}
\resizebox{0.75\linewidth}{!}{%
\begin{tabular}{ccc}
\toprule[0.75pt]
\begin{tabular}[c]{@{}c@{}}Auto-Interp.\\ Method\end{tabular} & \begin{tabular}[c]{@{}c@{}}Interp. Acc.\\ (IoU)\end{tabular} & Run Time       \\ \midrule[0.5pt]
Qwen3.5-VL-9B                                                 & 0.438 $\pm$ 0.249                                            & 1h 40min       \\
Ours                                                          & 0.432 $\pm$ 0.215                                            & \textbf{5 min} \\ \bottomrule[0.75pt]
\end{tabular}
}
\label{tab:compare_mllm}
\end{table}

\begin{figure*}[t]
  \centering
   \includegraphics[width=0.9\linewidth]{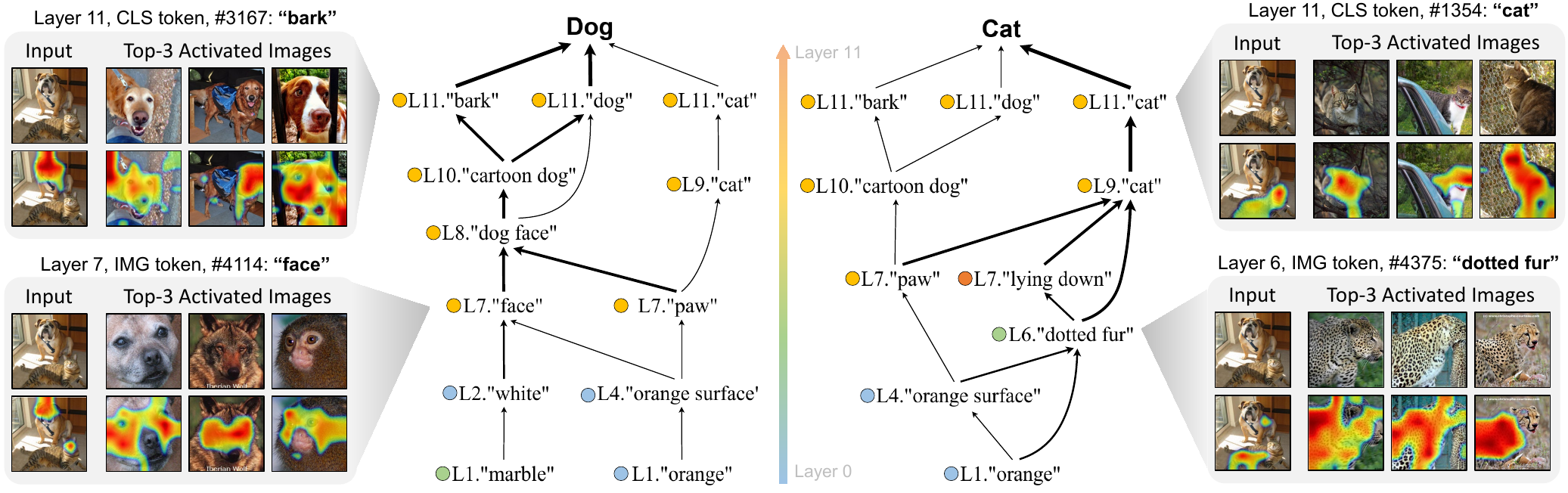}
   \caption{
   Visualization of concept circuits.
   For an input image containing both a dog and a cat, our method traces the unique causal pathways leading to each prediction.
   The circuit for ``dog'' composes primitive and intermediate concepts ({\it e.g.,} ``orange'' and ``marble'') into high-level semantics ({\it e.g.,} ``bark'' and ``dog'').
   In contrast, the circuit for ``cat'' relies on a different set of concepts ({\it e.g.,} ``dotted fur'').
   Our method can faithfully audit the inner workings of ViT and highlight responsible concepts.
   }
   \label{fig:information_flow}
\end{figure*}
\begin{figure*}[t]
  \centering
   \includegraphics[width=0.9\linewidth]{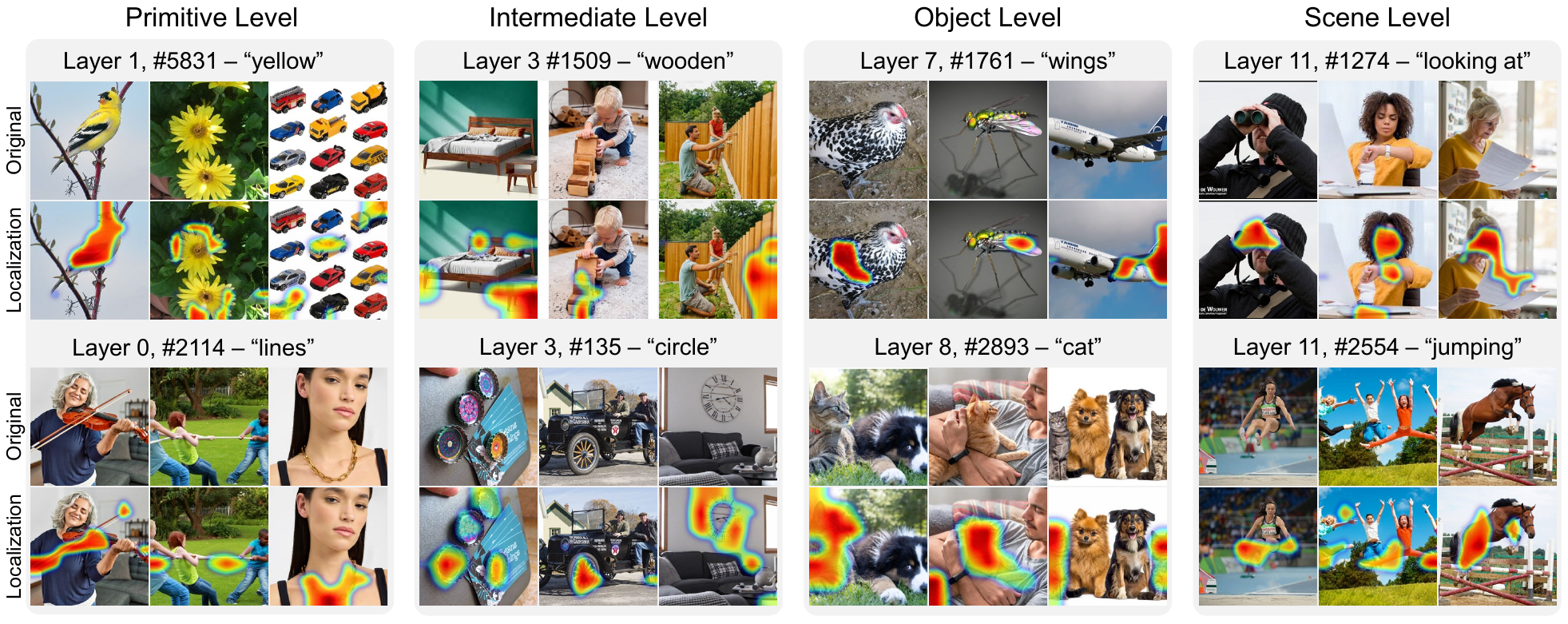}
   \caption{
   Localize concepts in the pixel space.
   Notably, our method can even localize highly abstract semantics, such as ``looking at'', by highlighting both the subject ({\it i.e.,} the person) and the object involved ({\it i.e.,} the paper).
   }
   \label{fig:localization}
\end{figure*}


\subsection{Evaluations of Interpretation Accuracy}
\label{subsec:interp_acc}

We develop a new metric based on our probing suite to evaluate the interpretation accuracy of SAE training data and concept sets for auto-interpretation.

\textbf{Settings:}
Similar to Sec.~\ref{subsec:benchmark}, we train SAEs on the representations (CLS and image tokens separately) extracted from the residual stream of each layer in {\it CLIP-ViT-B-32.}
(1) We compare different training data (downsampled to 60K), including our probing images, ImageNet~\citep{deng2009imagenet}, and MSCOCO~\citep{lin2014microsoft}.
(2) We compare using different concept sets to interpret the same SAEs, including our concept set ($\sim$16K), LAION frequent words (15K)~\citep{bhalla2024interpreting}, and Google Books common English words (20K)~\citep{oikarinen2022clip}.
(3) We compare using existing fine-grained interpretability datasets to train and interpret SAEs with our probing suite, including Broden~\citep{bau2017network} and LaBo~\citep{yang2023language}.
Note that LaBo only provides concept sets, so here we use ImageNet to train the SAEs and use LaBo's concepts for ImageNet to interpret the SAE features (Sec.~\ref{sec:method:tracing}).

\textbf{Metrics:}
We split the probing images into train and test sets (60K/4K).
Since ground-truth concepts are available for our test images (Sec.~\ref{subsec:probing}), we calculate {\it interpretation accuracy} by measuring the fraction of ground-truth concepts covered within the top-$K$ concepts read by SAEs from all layers.
We use CLIP-based semantic match rather than a raw string match to mitigate vocabulary circularity.

\textbf{Results:}
As shown in Tab.~\ref{tab:interpret_acc}, in the top-30 extracted concepts, SAEs trained on our probing images consistently outperform existing datasets by 2.9\% and 3.4\%.
Moreover, for the same SAEs, using our concept set consistently outperforms existing concept sets by 28.7\% and 37.4\%.
\new{Our probing suite outperforms existing fine-grained interpretability datasets by 27.3\%.}

\textbf{Compared with MLLM-based summarization:}
We compare our auto-interpretation method against an MLLM-based baseline~\citep{zhang2025large}. We implement it by using Qwen3.5-VL-9B to explain SAE features based on top-activating images. In their evaluation protocol, each interpreted concept label of the SAE feature will be grounded by GroundingDINO-SAM~\citep{liu2024grounding} on ImageNet-val images, and its spatial agreement with the SAE feature activation map will be evaluated using IoU. As shown in Tab.~\ref{tab:compare_mllm}, our interpretation achieves similar accuracy and is more stable, while requiring 20$\times$ less runtime per layer.

\subsection{Auditing}
\label{subsec:auditing}
In this section, we demonstrate how to use our toolbox to audit model behavior.

\textbf{Trace decision-making processes.}
Fig.~\ref{fig:information_flow} shows the concept circuit examples traced by our method.
Beyond faithfully identifying decision pathways, the circuits reveal a layer-wise progression that is similar to the human visual system:
early layers detect low-level primitives (colors, textures), while deeper layers compose these cues into higher-level semantics (objects, relations/motion).

\textbf{Localize concepts on pixels.}
Qualitatively, Figs.~\ref{fig:information_flow}~\&~\ref{fig:localization} show examples of our concept localizations.
Our method can accurately localize concepts across visual abstraction levels.
\new{Note that we do not manually choose the layer; the SAE activations determine it. 
For example, if an image strongly activates an SAE feature labeled ``wooden texture'' in its layer-3 representation, we use that layer-3 feature for localization.}
Quantitatively, as shown in Tab.~\ref{tab:localization}, our heatmaps on the Quantus~\citep{hedstrom2023quantus} benchmark improve over the existing attribution-based method~\citep{chefer2021generic} by 3.7\% using the VOC2007 dataset in terms of localization accuracy.

\begin{figure}[t]
  \begin{center}
  \includegraphics[width=1.0\linewidth]{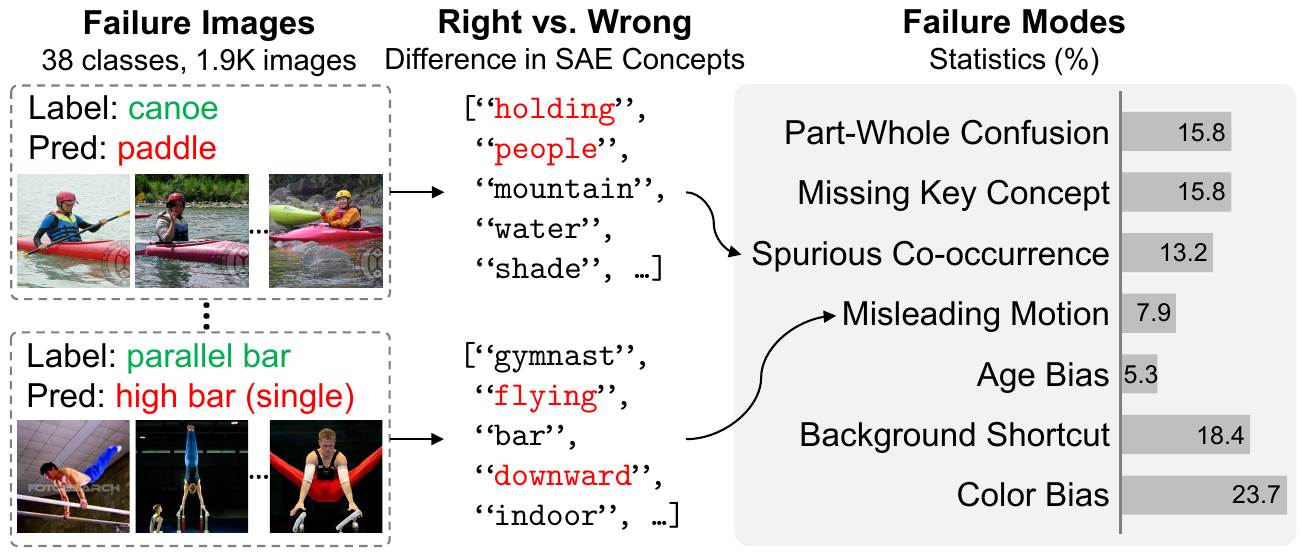}
  \end{center}
  \vspace{-8pt}
  \caption{
  Failure mode analysis.
  We identify seven failure modes of CLIP on the ImageNet-val set.
  For example, CLIP tends to misclassify parallel bar images as high bars when the gymnast is ``flying'' or oriented ``downward''.
  }
  \label{fig:failure_mode}
\end{figure}
\begin{table}
\centering
\caption{\new{Comparison of heatmap localization accuracy.}}
\resizebox{0.6\linewidth}{!}{%
\new{\begin{tabular}{lcc}
\toprule[0.75pt]
Method        & \begin{tabular}[c]{@{}c@{}}Point\\ Game\end{tabular} & \begin{tabular}[c]{@{}c@{}}Attribution\\ Localization\end{tabular} \\ \midrule[0.5pt]
Chefer et al. & 41.9                                                 & 32.3                                                               \\
Ours          & \textbf{45.0}                                        & \textbf{36.0}                                                      \\ \bottomrule[0.75pt]
\end{tabular}}
}
\label{tab:localization}
\end{table}

\textbf{Diagnose failure modes.}
Beyond instance-level auditing, understanding the failure patterns of the model at a global level is essential for improving its robustness.
To this end, we demonstrate how to leverage our {\it ViSAE} to diagnose the failure modes of the {\it CLIP-ViT-B-32} model on the ImageNet validation set.
Specifically, we identify all 38 classes for which more than 40\% of the images are consistently misclassified into the same incorrect class.
For each such class, we use our trained SAEs to extract the most frequently activated concepts from both correctly and incorrectly classified images.
By comparing these two concept groups, we summarize key semantic differences, eventually organizing them into seven distinct failure modes (Fig.~\ref{fig:failure_mode}).
Building on these analyses, our {\it ViSAE} can be extended to support data quality control by identifying mislabeled, ambiguous, or systematically biased samples.

\begin{table}[t]
\centering
\caption{Model steering accuracy on the WaterBird dataset. The Worst Group refers to ``Waterbird on Land''.
}
\resizebox{0.95\linewidth}{!}{%
\begin{tabular}{lcccc}
\toprule[1pt]
\multicolumn{1}{c}{Method} & \begin{tabular}[c]{@{}c@{}}Steer\\ Spuri. Corr.\end{tabular} & \begin{tabular}[c]{@{}c@{}}Overall\\ Acc. (\%)\end{tabular} & \begin{tabular}[c]{@{}c@{}}Worst Group\\ Acc. (\%)\end{tabular} & $\Delta$                          \\ \midrule[0.75pt]
\multirow{2}{*}{CBM}       & None                                                         & -                                                           & 37.3                                                            & -                                 \\
                           & Remove                                                       & -                                                           & 51.8                                                            & {\color{deepgreen} $+$ 14.5}          \\ \midrule[0.5pt]
\multirow{2}{*}{SpLiCE}    & None                                                         & -                                                           & 48.0                                                            & -                                 \\
                           & Remove                                                       & -                                                           & 60.0                                                            & {\color{deepgreen} $+$ 12.0}          \\ \midrule[0.5pt]
\new{\multirow{2}{*}{DN-CBM}}    & \new{None}                                                         & \new{-}                                                           & \new{57.5}                                                            & \new{-}                                 \\
                           & \new{Remove}                                                       & \new{-}                                                           & \new{71.3}                                                            & 
                           {\color{deepgreen} $+$ 13.8}          \\ \midrule[0.5pt]
\new{\multirow{2}{*}{PCBM} }     & \new{None}                                                         & \new{-}                                                           & \new{50.3}                                                            & \new{-}                                 \\
                           & \new{Remove}                                                       & \new{-}                                                           & \new{74.7}                                                            & {\color{deepgreen} $+$ 24.4}          \\ \midrule[0.5pt]
\multirow{2}{*}{Joseph et al.}    & None                                                         & 68.8                                                           & 22.4                                                            & -                                 \\
                           & Remove                                                       & 71.9                                                           & 24.6                                                            & {\color{deepgreen} $+$ 2.2}          \\ \midrule[0.5pt]
\multirow{2}{*}{COAR}    & None                                                         & 88.0                                                           & 64.0                                                            & -                                 \\
                           & Remove                                                       & 91.0                                                           & 83.0                                                            & {\color{deepgreen} $+$ 19.0}          \\ \midrule[0.5pt]
\multirow{3}{*}{Ours}      & None                                                         & 79.7                                                        & 50.3                                                            & -                                 \\
                           & Enhance                                                      & 74.5                                                        & 5.3                                                             & {\color{red} $-$ 45.0}                \\
                           \rowcolor{Gray}& Remove                                                       & 85.2                                                        & \textbf{98.5}                                                   & \textbf{\color{deepgreen} $+$ 48.2} \\ \bottomrule[1pt] 
\end{tabular}}
\label{tab:steering}
\end{table}%


\subsection{Steering}
\label{subsec:steering}

In this section, we demonstrate the capability of our {\it ViSAE} in steering model behavior.

\textbf{Settings.}
We use the WaterBirds dataset~\citep{sagawadistributionally} to evaluate robustness against spurious correlations between bird species (land/water) and their backgrounds.
The training set amplifies this spurious correlation ({\it e.g.,} waterbirds on water), while a 5\% ``worst-group'' in the test set breaks it ({\it e.g.,} waterbirds on land).
A linear classifier trained on {\it CLIP-ViT-B-32}'s final-layer CLS tokens achieves only 50.3\% accuracy on this worst group, confirming heavy reliance on background cues.
\new{We compare with Concept Bottleneck Models (CBM)~\citep{koh2020concept},  SpLiCE~\citep{bhalla2024interpreting}, DN-CBM~\citep{rao2024discover}, PCBM~\citep{yuksekgonulpost}, Joseph et al.~\citep{joseph2025steering}, and COAR~\citep{shah2024decomposing} that can steer the model.}

\textbf{Edit spurious concepts.}
We mitigate spurious correlations by using our SAE to ablate background-related concepts (e.g., ``grass'', ``land'') in the worst-group samples.
This is done by setting their activations to zero, reconstructing new CLS tokens, and re-classifying.
As shown in Tab.~\ref{tab:steering}, this intervention boosts worst-group accuracy by 48.2\%.
Conversely, enhancing these concepts degrades performance by 45.0\%, demonstrating precise bidirectional control via our concept-level ``knobs''.

\textbf{Ablation study on different probing image sets.}
We further conduct ablation studies to evaluate the influence of the probing image set for training SAEs on steering performance.
To ensure a fair comparison, we keep the SAE architecture and all training hyperparameters fixed, and train SAEs using ImageNet, MSCOCO, and our curated probing image set, respectively.
As shown in Tab.~\ref{tab:ablation}, the SAEs trained on our probing image set significantly outperform their counterparts, even though those also achieve improvements over existing methods.

\begin{table}  
\centering
\caption{
    Ablation study on the impact of probing image sets for model steering.
}
\resizebox{0.65\columnwidth}{!}{%
\begin{tabular}{ccc}
\toprule[0.75pt]
\begin{tabular}[c]{@{}c@{}}SAE\\ Probing Set\end{tabular} & \begin{tabular}[c]{@{}c@{}}Steer\\ Spuri. Corr.\end{tabular} & \begin{tabular}[c]{@{}c@{}}Worst Group\\ Acc. (\%)\end{tabular} \\ \midrule[0.5pt]
None                                                         & None                                                         & 50.3                                                            \\ \midrule[0.4pt]  
ImageNet                                                  & Remove                                                       & 63.9                                                      \\
MSCOCO                                                    & Remove                                                       & 95.4                                                            \\
\rowcolor{Gray} Ours                                         & Remove                                                       & \textbf{98.5}                                                   \\ \bottomrule[0.75pt]
\end{tabular}
}
\label{tab:ablation}
\end{table}



\section{Related Work}
\label{sec:related_work}

In this section, we will discuss the most related works. A more detailed version is in \textbf{Appendix~\ref{appendix:related_work}.}

\textbf{Interpretable Machine Learning (IML).}
Existing IML methods 
are broadly categorized into two paradigms.
{\it Post-hoc methods}, such as GradCAM~\citep{selvaraju2017grad}, LIME~\citep{ribeiro2016should}, and SHAP~\citep{lundberg2017unified}, typically provide a saliency map of input pixels;
{\it Intrinsic methods,} such as ProtoPNet~\citep{chen2019looks} and explanation-guided learning~\citep{ross2017right}, incorporate interpretability directly into the model architecture.
However, the former is limited to input-output correlations, and the latter relies on custom architectures that are not easily generalizable across tasks
~\citep{rudin2019stop, wang2025beyond}.
Although there are existing attempts using interpretations as guidance to provide additional supervision for the model training~\citep{li2023data,li2024deal,li2024beyond,ma2025there}, the inner workings of the model remain a black box.
Differently, our method interprets internal representations post hoc, without requiring architectural changes to the explainee model.

\textbf{Concept-based interpretability.}
Concept-based methods address the limited intelligibility of saliency maps by explaining predictions via human-understandable concepts. Techniques include TCAV~\citep{kim2018interpretability}
, Network Dissection~\citep{bau2017network} 
, and Concept Bottleneck Models~\citep{koh2020concept}. 
There are also existing attempts to use predefined concept bottlenecks for failure detection~\citep{nguyen2025interpretable}.
However, these approaches rely on predefined concept sets or annotations, limiting their scalability in open-world settings~\citep{yuksekgonulpost, margeloiu2021concept}.
Different from existing works, we use SAEs to ``read'' concepts directly from model representations without supervised concept labels.

\textbf{Mechanistic Interpretation (MI).}
MI methods aim to reverse-engineer the internal mechanisms of deep models~\citep{nandaprogress, bereskamechanistic}.
{\it Bottom-up} approaches, such as the Circuits framework~\citep{olah2020zoom, conmy2023towards}, dissect neural connectivity but yield low-level graphs that lack human interpretability~\citep{marks2024sparse, peng2026inside}.
{\it Top-down} methods like SAEs~\citep{olshausen1997sparse, ng2011sparse} learn disentangled features to address superposition. In language models, SAEs are trained on large corpora and interpreted via LLM summarization~\citep{bills2023language}. For vision, however, SAEs face two key challenges: biased datasets with limited concept coverage, and feature interpretation remains subjective~\citep{thasarathan2025universal, pach2025sparse}.
Our toolbox bridges these gaps by curating data and auto-interpretation.
\section{Conclusion}
Different from existing works that primarily focus on improving the architecture design of SAE, we improve the interpretability from a data perspective.
Motivated by neuroscience, we develop a compact, diagnostic toolbox, ViSAE, that enables SAEs to efficiently interpret the inner workings of ViTs.
Specifically, our toolbox (data and algorithm) enables automatic SAE feature interpretation at scale and faithful concept circuit tracing.
We show that {\it ViSAE} is not only a powerful auditing tool for identifying spurious correlations and failure modes but also enables effective model steering through concept-level interventions.


\textbf{Limitations.}
ViSAE inherits several limitations from SAE-based mechanistic interpretation. First, although our probing suite improves concept coverage and automatic interpretation, SAE features may still exhibit feature absorption or feature composition: a single SAE feature can absorb multiple correlated visual concepts, while a high-level semantic concept may be compositionally represented by multiple SAE features across layers. Therefore, the resulting concept labels and circuit edges should be interpreted as useful approximations of model mechanisms, rather than a guaranteed one-to-one mapping between features and human concepts. Second, our automatic concept reading depends on the coverage of the concept vocabulary and the vision-language embedding space used for matching, which may miss rare, abstract, or domain-specific semantics. Third, our empirical evaluation focuses primarily on CLIP-style ViTs and natural images. Extending ViSAE to other architectures, larger vision-language models, and specialized domains remains an important direction for future work.

\section*{Acknowledgement}
This work is supported by the National Science Foundation under grant numbers CAREER 2340074, SLES 2416937, and III CORE 2412675, the National Institutes of Health under grant number R21CA301093, and the Department of Defense under grant number AFOSR FA9550-23-1-0494.
Any opinions, findings, and conclusions or recommendations expressed in this material are those of the authors and do not reflect the views of the supporting entities.




\section*{Impact Statement}


This paper presents work whose goal is to advance the field of Machine
Learning. There are many potential societal consequences of our work, none of
which we feel must be specifically highlighted here.



\bibliography{icml2026_conference}
\bibliographystyle{icml2026}

\newpage
\appendix
\onecolumn
\section*{Appendix}


\section{Related Work}
\label{appendix:related_work}

\textbf{Interpretable Machine Learning (IML).}
IML methods aim to uncover the reasons behind model predictions and are typically categorized as:
Post-hoc methods, such as GradCAM~\citep{selvaraju2017grad}, LIME~\citep{ribeiro2016should}, and SHAP~\citep{lundberg2017unified}, provide explanations by attributing predictions to input features;
Intrinsic methods, such as ProtoPNet~\citep{chen2019looks} and explanation-guided learning~\citep{ross2017right}, incorporate interpretability directly into the model architecture by design.
However, the former is often limited to surface-level input-output relationships, while the latter depends on custom architectures that are not easily generalizable across tasks~\citep{adebayo2018sanity, rudin2019stop}.
In contrast, our method decouples interpretation from prediction and instead analyzes internal representations post hoc, without requiring architectural changes to the explainee model.

\textbf{Concept-based interpretability.}
Concept-based methods emerged as a response to the limitations of attribution methods, where saliency maps often fail to provide human-interpretable explanations~\citep{adebayo2018sanity, kindermans2019reliability}.
TCAV~\citep{kim2018interpretability} uses curated probing sets to evaluate a model’s sensitivity to predefined concept directions.
Network Dissection~\citep{bau2017network} assigns semantics to individual neurons using human-annotated labels.
ACE automatically discovers salient concept clusters in latent space.
Concept Bottleneck Models (CBMs)~\citep{koh2020concept} enforce a human-defined concept layer within the network, enabling transparency and intervention.
However, these methods typically require concept annotations or assume a closed-world setting with a fixed concept vocabulary, making them struggle to scale up to open-world concept discovery that does not assume the set of concepts is a known prior~\citep{yuksekgonulpost, margeloiu2021concept}.
Our method differs by directly extracting concepts from pretrained models without requiring concept supervision or architecture changes.

\textbf{Mechanistic Interpretation (MI).}
MI methods~\citep{nandaprogress, bereskamechanistic} aim to uncover the internal computational mechanisms of deep models and have shown promising progress, particularly in language models.
Bottom-up approaches, such as the Circuits framework~\citep{olah2020zoom, conmy2023towards}, dissect individual neurons and their connectivity to reveal functional subcomputations.
However, the resulting units ({\it e.g.,} neurons) are often not interpretable to humans~\citep{marks2024sparse}.
Top-down approaches, including representation engineering~\citep{zou2023representation} and Sparse Autoencoders (SAEs)~\citep{olshausen1997sparse, ng2011sparse}, address this by learning disentangled, monosemantic features that map more naturally to human-understandable concepts, mitigating the feature superposition issue.
While SAEs decompose polysemantic representations into monosemantic features, ensuring comprehensive coverage and interpreting their semantic meanings remains challenging.
For language models, existing approaches~\citep{huben2023sparse, lieberum2024gemma} typically train SAEs on massive text corpora, such as the Pile ($\sim$7M)~\citep{pile} or Gemma ($\sim$3T)~\citep{team2024gemma}, to broaden concept coverage, and interpret features by prompting LLMs ({\it e.g.}, GPT-4~\citep{achiam2023gpt}) to summarize the semantics of top-activating examples~\citep{bills2023language}.
For vision models, however, available datasets are usually biased toward object-level concepts ({\it e.g.,} ImageNet~\citep{deng2009imagenet}), might not cover the full spectrum of visual processing.
Furthermore, the top-activating images of SAE features often show ambiguous semantics, making their interpretation subjective~\citep{thasarathan2025universal, pach2025sparse}.

\section{Concept Coverage Calculation}
\label{appendix:concept_coverage}
To measure concept coverage in a dataset-agnostic manner, we leverage a Top Percentile Method that evaluates how well each concept is represented by the most similar images in the dataset.
For each concept $c_i$, we compute the cosine similarity between the concept's text embedding and all image embeddings (CLIP-ViT-B-32) in the dataset, yielding a similarity vector $\mathbf{s}_i \in \mathbb{R}^N$ where $N$ is the dataset size.
Rather than relying on maximum similarity (which can be noisy) or overall mean similarity (which may be dominated by irrelevant images), we calculate the mean similarity of the top $k$ most similar images, where $k = \mathrm{max} \; (1, [N\cdot p/100])$ and $p$ is a small percentile (typically 0.005\%), namely:
\begin{equation}
    \mathrm {Coverage} \; \mathrm {Score} \; (c_i) = \frac{1}{k} \sum_{j=1}^{k} s_{i,\mathrm{top}-j},
\end{equation}
where $s_{i,\mathrm{top}-j}$ represents the $j$-th highest similarity score between concept $c_i$ and the images in the dataset.
A concept is considered ``covered'' at threshold $\tau$ if $\mathrm {Coverage} \; \mathrm {Score} \; (c_i) \ge \tau$.
In practice, we set $\tau = 0.25$ as a meaningful similarity between modalities.

This approach is robust to dataset size variations and provides a stable measure of concept representation quality by focusing on the images that most strongly exhibit each concept, while avoiding the influence of outliers or the vast majority of irrelevant images.

\section{Concept Count Calculation}
\label{appendix:concept_count}
To understand the semantic distribution of existing concept sets across different abstraction levels, we perform a match analysis that assigns each of their concepts to its best-matching abstraction level in our ground truth taxonomy.
Given an existing concept set $\mathcal{C} = \{c_1, c_2, ..., c_N\}$ and ground truth concepts organized by abstraction levels $\mathcal{G}_\mathrm{gt}  = \{\mathcal{G}_\mathrm{primitive}, \mathcal{G}_\mathrm{intermediate}, \mathcal{G}_\mathrm{object}, \mathcal{G}_\mathrm{scene} \}$, we first compute the semantic similarity between each new concept and all ground truth concepts $\mathrm{sim}(c_i, g_i) $ using CLIP text embeddings (CLIP-ViT-B-32).
We then identify the best-matching ground truth concept for each new concept:
\begin{equation}
    g_i^{*} = \operatorname*{arg\,max}_{g_j \in \bigcup_{l} g_l}\; \mathrm{sim}(c_i, g_j)
\end{equation}
A concept $c_i$ is considered well-matched if its maximum similarity exceeds a threshold $\tau$ (0.9 in practice).
Each well-matched concept is then assigned to the abstraction level of its best-matching ground truth concept.

This analysis reveals the semantic composition of existing concept sets, showing how many concepts align with each abstraction level (primitive, intermediate, object, scene) and identifying concepts that may not represent visual concepts.

\section{Details on SAE Architectures}
\label{appendix:SAE}
In this section, we provide details for the SAE variants used in our study. These variants differ primarily in how they induce sparsity in the hidden representation $\mathbf{h}$.

\textbf{ReLU SAE}~\citep{bricken2023towards}.
This is the standard sparse autoencoder using ReLU nonlinearity followed by an $L_1$ sparsity penalty. Given the input representation $\mathbf{x}\in \mathbb{R}^{d_\mathrm{in}}$, encoder, decoder weights $\mathbf{W}_\mathrm{enc}, \mathbf{W}_\mathrm{dec}^{\top} \in  \mathbb{R}^{d_\mathrm{hid}\times d_\mathrm{in}}$ and biases $\mathbf{b}_\mathrm{enc},\mathbf{b}_\mathrm{dec} \in \mathbb{R}^{d_\mathrm{hid}}$, the hidden representation $\mathbf{h}\in \mathbb{R}^{d_\mathrm{hid}}$ is given by:
\begin{equation}
\mathbf{h} = \mathrm{ReLU}(\mathbf{W}_\mathrm{enc} \mathbf{x} + \mathbf{b}_\mathrm{enc}),
\end{equation}
and the reconstruction is:
\begin{equation}
\hat{\mathbf{x}} = \mathbf{W}_\mathrm{dec} \mathbf{h} + \mathbf{b}_\mathrm{dec}.
\end{equation}
The ReLU SAE is trained to minimize a loss function:
\begin{equation}
\mathcal{L} = \mathcal{L}_\text{reconstruction} + \mathcal{L}_\text{sparsity} = \|\mathbf{x} - \hat{\mathbf{x}}\|_2^2 + \lambda \|\mathbf{h}\|_1,
\end{equation}
where $\lambda$ is the regularization coefficient to control sparsity.
This variant is simple and effective, but sparsity is indirectly controlled by $\lambda$.

\textbf{BatchTopK SAE}~\citep{bussmann2024batchtopk}.
Instead of applying a soft sparsity penalty, this variant enforces hard sparsity by retaining only the top-$k$ activations across a batch of $n$ samples. Specifically, it retains the $n \times k$ largest activations across the batch and zeros out all others:
\begin{equation}
\mathbf{h} = \mathrm{BatchTopK}(\mathbf{W}_\mathrm{enc} \mathbf{x} + \mathbf{b}_\mathrm{enc}).
\end{equation}
The loss becomes:
\begin{equation}
\mathcal{L} = \|\mathbf{x} - \mathbf{W}_\mathrm{dec} \mathbf{h} + \mathbf{b}_\mathrm{dec}\|_2^2.
\end{equation}
This provides deterministic sparsity and is particularly suitable for interpretability-focused applications.

\textbf{Matryoshka SAE}~\citep{bussmann2025learning}.
Matryoshka SAE is inspired by the idea of nested sparsity levels. It produces multiple nested representations $\{\mathbf{h}^{(1)}, \ldots, \mathbf{h}^{(K)}\}$ such that each $\mathbf{h}^{(k)}$ satisfies $\mathbf{h}^{(1)} \subseteq \ldots \subseteq \mathbf{h}^{(K)}$, where $\mathbf{h}^{(K)}=\mathrm{BatchTopK}(\mathbf{W}_\mathrm{enc} \mathbf{x} + \mathbf{b}_\mathrm{enc})$. The reconstruction loss is computed over all levels:
\begin{equation}
\mathcal{L} = \sum_{k=1}^K \|\mathbf{x} - \mathbf{W}_\mathrm{dec} \mathbf{h}^{(k)} + \mathbf{b}_\mathrm{dec}\|_2^2.
\end{equation}
This encourages a hierarchical structure in the learned features and facilitates interpretability at multiple granularity levels.

\textbf{JumpReLU SAE}~\citep{rajamanoharan2024jumping}.
JumpReLU replaces the standard ReLU with a modified activation function that enforces a minimum activation threshold:
\begin{equation}
\mathrm{JumpReLU}(z) = \begin{cases}
z, & z > \tau \\
0, & \text{otherwise}
\end{cases},
\end{equation}
where $\tau$ is a fixed threshold (e.g., $\tau=0.001$). This nonlinearity encourages fewer active units by cutting off low activations more aggressively than ReLU, resulting in sparser codes even without explicit sparsity penalties.

\textbf{Gated SAE}~\citep{rajamanoharan2024improving}.
Gated SAE uses multiplicative gating to modulate activations. Each hidden unit has a learned gate $g_i \in [0, 1]$, typically computed via a sigmoid:
\begin{equation}
\mathbf{h}_i = \sigma(\mathbf{a}_i^\top \mathbf{x}) \cdot \mathrm{ReLU}(\mathbf{w}_i^\top \mathbf{x} + b_i).
\end{equation}
This allows the model to selectively suppress irrelevant features and provides an adaptive mechanism to control sparsity, potentially improving both interpretability and flexibility.

\section{Evaluation Metrics}
\label{appendix:evaluation_metircs}

\textbf{Reconstruction Error.}
This metric quantifies how well the autoencoder reconstructs the input data from the sparse code. Formally, given input $\mathbf{x} \in \mathbb{R}^{d_\mathrm{in}}$ and reconstructed output $\hat{\mathbf{x}} = \mathbf{W}_\mathrm{dec} \mathbf{h} + \mathbf{b}_\mathrm{dec}$, the Reconstruction Error is defined as:
\begin{equation}
\mathrm{Reconstruction\ Error} = \mathbb{E}_{\mathbf{x} \sim \mathcal{D}} \left[\|\mathbf{x} - \hat{\mathbf{x}}\|_2^2 \right].
\end{equation}
Lower Reconstruction Error indicates that the SAE preserves more input information. However, extremely low reconstruction error may come at the cost of losing sparsity or interpretability.

\textbf{Monosemanticity}~\citep{pach2025sparse}.
Monosemanticity is a measure of how consistently a neuron responds to semantically similar inputs. Intuitively, a neuron is considered \emph{monosemantic} if its highest activations occur for a group of inputs that are semantically coherent (e.g., images depicting the same object or concept). Specifically, it measures the visual similarity between the top-$k$ activated inputs for each neuron.

Formally, let $f(\mathbf{x}) = \mathbf{h} \in \mathbb{R}^{d_\mathrm{hid}}$ be the encoder output for input $\mathbf{x}$, and let $h_i$ denote the activation of the $i$-th neuron. For each neuron $i \in \{1, \dots, d_\mathrm{hid}\}$, we identify the top-$k$ inputs from the dataset $\mathcal{D}$ that elicit the highest activations:
\begin{equation}
\mathcal{T}_i = \mathrm{TopK}\left( \{ (\mathbf{x}, h_i(\mathbf{x})) \mid \mathbf{x} \in \mathcal{D} \} \right).
\end{equation}
We then compute the average pairwise cosine similarity between the embeddings of the top-$k$ input images. Let $\phi(\mathbf{x}) \in \mathbb{R}^d$ be a feature embedding of image $\mathbf{x}$ obtained from the CLIP ViT-B/32 model. The monosemanticity score for neuron $i$ is:
\begin{equation}
\mathrm{Mono}(i) = \frac{2}{k(k-1)} \sum_{1 \le p < q \le k} \frac{\phi(\mathbf{x}_p)^\top \phi(\mathbf{x}_q)}{\|\phi(\mathbf{x}_p)\|_2 \cdot \|\phi(\mathbf{x}_q)\|_2},
\quad \text{where } \{\mathbf{x}_1, \dots, \mathbf{x}_k\} = \mathcal{T}_i.
\end{equation}
Finally, the overall monosemanticity score for the autoencoder is obtained by averaging over all neurons:
\begin{equation}
\mathrm{Monosemanticity} = \frac{1}{d_\mathrm{hid}} \sum_{i=1}^{d_\mathrm{hid}} \mathrm{Mono}(i).
\end{equation}
Higher values indicate that neurons respond selectively to visually similar inputs, which supports more interpretable and disentangled representations.

\textbf{Decoder Orthogonality}~\citep{zaigrajew2025interpreting}.
To enhance interpretability, it is desirable that the learned basis features (\textit{i.e.}, decoder columns) are disentangled. One way to promote this is to encourage orthogonality among decoder vectors. Let $\mathbf{W}_\mathrm{dec} = [\mathbf{w}_1, \dots, \mathbf{w}_{d_\mathrm{hid}}] \in \mathbb{R}^{d_\mathrm{in} \times d_\mathrm{hid}}$ denote the decoder weight matrix. The Decoder Orthogonality is defined as the mean pair-wise cosine similarity between each pair of decoder columns:
\begin{equation}
\mathrm{Decoder\ Orthogonality} = \frac{2}{d_\mathrm{hid}(d_\mathrm{hid} - 1)} \sum_{1 \le i < j \le d_\mathrm{hid}} \mathbf{w}_i^\top \mathbf{w}_j.
\end{equation}
A smaller value implies greater orthogonality and lower redundancy among the learned features. Perfect orthogonality occurs when all decoder vectors are mutually orthogonal unit vectors.

\textbf{Dead Neuron.}
This metric measures the fraction of hidden units that are never activated across a dataset. A hidden neuron is considered ''dead`` if its activation is zero for all inputs in a dataset $\mathcal{D}$. Let $\mathbf{h}(\mathbf{x})$ be the hidden representation on input $\mathbf{x}$ and $h_i(\mathbf{x})$ the $i$-th dimension. Define the dead neuron set:
\begin{equation}
\mathcal{D}_\mathrm{dead} = \left\{ i \in \{1, \dots, d_\mathrm{hid}\} \;\middle|\; \sum_{\mathbf{x} \sim \mathcal{D}}h_i(\mathbf{x})=0 \right\}.
\end{equation}
The Dead Neuron ratio is:
\begin{equation}
\mathrm{Dead\ Neuron} = \frac{|\mathcal{D}_\mathrm{dead}|}{d_\mathrm{hid}}.
\end{equation}
A high dead neuron ratio indicates underutilization of the model capacity, which may suggest over-regularization or poor feature allocation. On the other hand, a moderate level of dead neurons may naturally emerge in highly sparse encoders.

\section{Implementation Details}
\label{appendix:implementation_details}
\textbf{Training Details.}
We train all Sparse Autoencoders (SAEs) on our probing image set using the \textit{cls} tokens and \textit{image} tokens from the residual stream of each layer in the CLIP ViT-B/32 model (each layer has two SAEs).
Each SAE consists of an overcomplete linear encoder and a sparse decoder, with the decoder columns constrained to unit $\ell_2$ norm. For benchmark experiment, we vary the expansion factor $ef \in \{2, 4, 8, 16, 32\}$, defined as $d_{\mathrm{hid}} = ef \cdot d_{\mathrm{in}}$, and $L_0$ sparsity $L_0\in \{8,16,32,64,128\}$ for all SAE architectures.

\textbf{Optimization and Scheduler.}
We train all models using a modified Adam optimizer that enforces unit-norm constraints on decoder columns. We use a fixed batch size of $4096$, learning rate $\eta = 3 \times 10^{-4}$, and train for $100$ epochs. No learning rate decay or warmup is applied.

\textbf{Hyperparameter Choice.}
We control sparsity in BatchTopK and Matryoshka SAEs by directly setting the top-$k$ values $k \in {8, 16, 32, 64, 128}$. For ReLU, JumpReLU, and Gated SAEs, we perform a sweep over sparsity regularization strength $\lambda$ to approximate the target $L_0$ sparsity levels. In Matryoshka SAEs, we use nested hidden representations with cumulative fractions ${\frac{1}{32}, \frac{1}{16}, \frac{1}{8}, \frac{1}{4}, \frac{1}{2}, 1}$. For JumpReLU SAEs, we fix the jump threshold $\tau = 0.001$ throughout all experiments. For the Monosemanticity metric, we compute pair-wise similarity between $k=9$ top activated images for each feature basis, for the Interpretation Accuracy metric, we retrieve $k=3$ concepts from the concept set for each feature basis.

\section{Additional Experimental Results}
\label{appendix:benchmark}

\begin{figure}[t]
  \centering
   \includegraphics[width=0.98\linewidth]{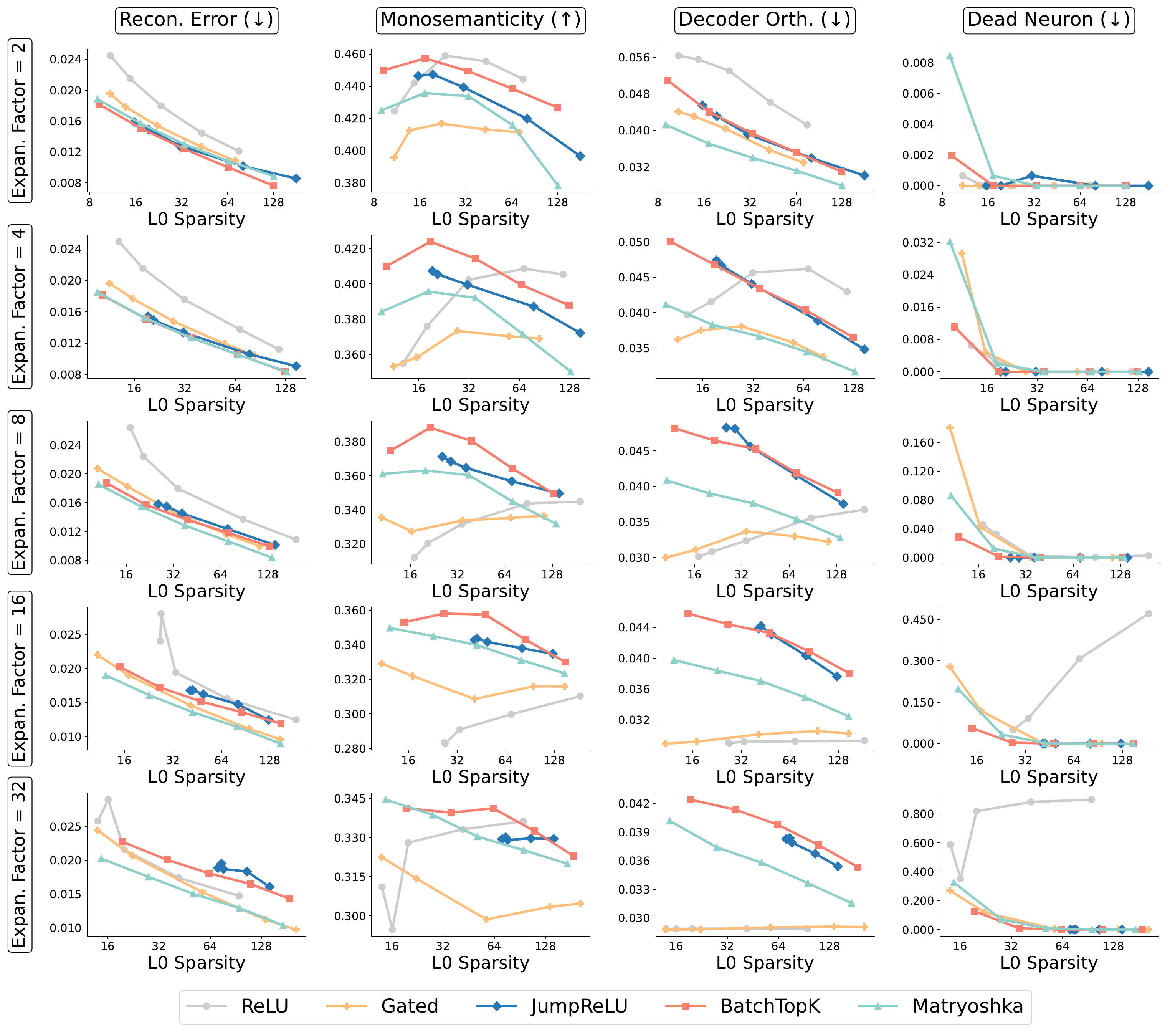}
   \caption{
   Full Benchmark results for {\it cls} tokens.
   }
   \label{fig:benchmark_full_cls}
\end{figure}
\begin{figure}[t]
  \centering
   \includegraphics[width=0.98\linewidth]{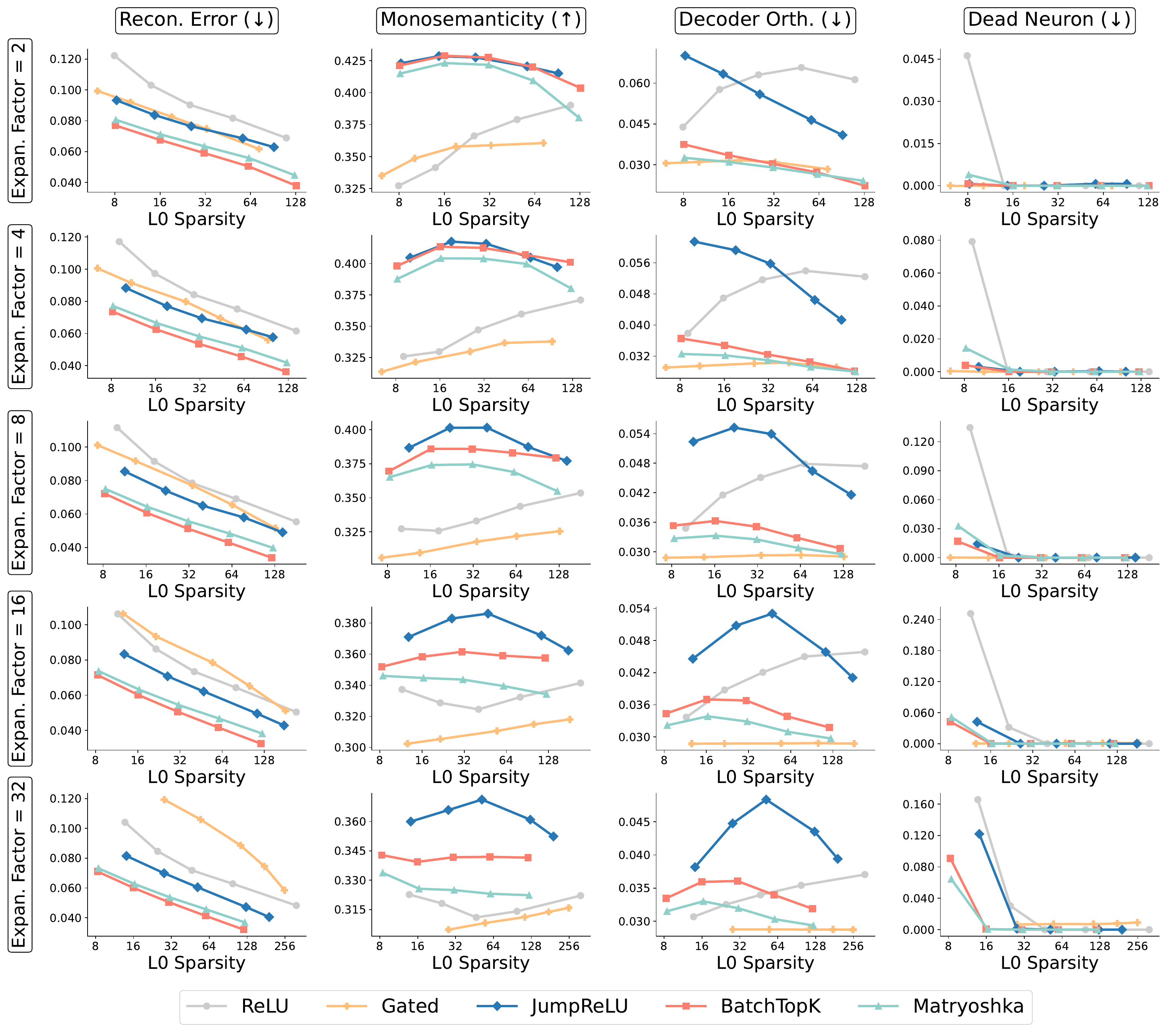}
   \caption{
   Full Benchmark results for image tokens.
   }
   \label{fig:benchmark_full_img}
\end{figure}

We provide full benchmark results in Figs.~\ref{fig:benchmark_full_cls}\&~\ref{fig:benchmark_full_img}.

\end{document}